\documentclass[letterpaper, 10 pt, conference]{ieeeconf}  

\IEEEoverridecommandlockouts                              

\usepackage{graphicx}
\usepackage{picinpar}
\usepackage{amsmath,bm}
\usepackage{url}
\usepackage{colortbl}
\usepackage{soul}
\usepackage{multirow}
\usepackage{pifont}
\usepackage{color}
\usepackage{alltt}
\usepackage[hidelinks]{hyperref}
\usepackage{enumerate}
\usepackage{siunitx}
\usepackage{breakurl}
\usepackage{epstopdf}
\usepackage{pbox}
\usepackage{indentfirst}
\usepackage{mathtools}
\usepackage{booktabs}
\usepackage{makecell}
\usepackage{subfigure}
\usepackage{float}
\usepackage{threeparttable}
\usepackage{algorithmic}
\usepackage{algorithm}
\usepackage{amsfonts}
\usepackage[allow-number-unit-breaks]{siunitx}
\usepackage[dvipsnames]{xcolor}
\usepackage{cite}

\newcommand{\etalcite}[1]{et al.~\cite{#1}}

\title{\LARGE \bf LIO-EKF: High Frequency LiDAR-Inertial Odometry\\ using Extended Kalman Filters}

\author{Yibin~Wu ~Tiziano~Guadagnino ~Louis~Wiesmann ~Lasse~Klingbeil ~Cyrill~Stachniss ~Heiner~Kuhlmann 
\thanks{All authors are with the Institute of Geodesy and Geoinformation, University of Bonn, Bonn, Germany. Cyrill Stachniss is additionally with the Department of Engineering Science at the University of Oxford, UK, and with the Lamarr Institute for Machine Learning and Artificial Intelligence, Germany. {\{{\tt\small firstname.lastname}\}{\tt\small @igg.uni-bonn.de}}.} 
\thanks{This work was funded by the Deutsche Forschungsgemeinschaft (DFG, German Research Foundation) under Germany's Excellence Strategy-EXC 2070-390732324.}%
}

\begin{document}

\maketitle
\begin{abstract}
Odometry estimation is crucial for every autonomous system requiring navigation in an unknown environment. In modern mobile robots, 3D LiDAR-inertial systems are often used for this task. By fusing LiDAR scans and IMU measurements, these systems can reduce the accumulated drift caused by sequentially registering individual LiDAR scans and provide a robust pose estimate. Although effective, LiDAR-inertial odometry systems require proper parameter tuning to be deployed. In this paper, we propose LIO-EKF, a tightly-coupled LiDAR-inertial odometry system based on point-to-point registration and the classical extended Kalman filter scheme. We propose an adaptive data association that considers the relative pose uncertainty, the map discretization errors, and the LiDAR noise. In this way, we can substantially reduce the parameters to tune for a given type of environment. The experimental evaluation suggests that the proposed system performs on par with the state-of-the-art LiDAR-inertial odometry pipelines but is significantly faster in computing the odometry. The source code of our implementation is publicly available
(https://github.com/YibinWu/LIO-EKF).
\end{abstract}
\IEEEpeerreviewmaketitle

\section{Introduction}

\begin{figure}
    \centering
    \subfigure[Handheld device on the campus of the University of Oxford.]{\includegraphics[width = 4cm]{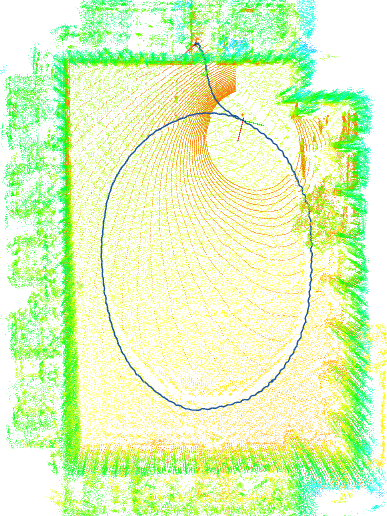}\label{ncd}}
    \quad
    \subfigure[Wheeled mobile robot on a university campus in Shanghai.]{\includegraphics[width = 4cm]{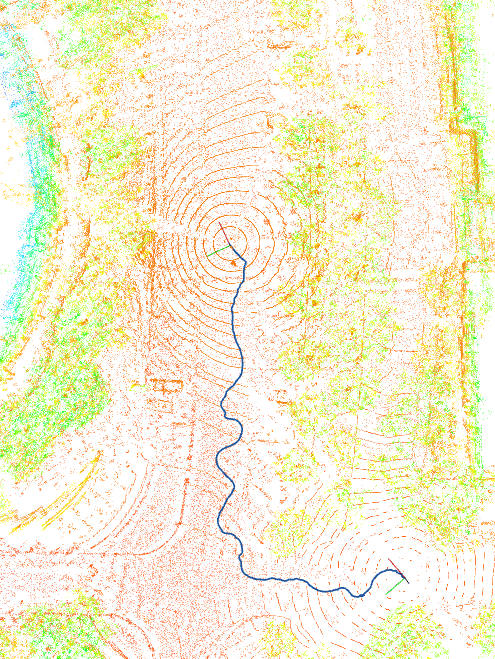}\label{m2dgr}}
    \quad
    \subfigure[Car in Hong Kong city center.]{\includegraphics[width = 8.5cm]{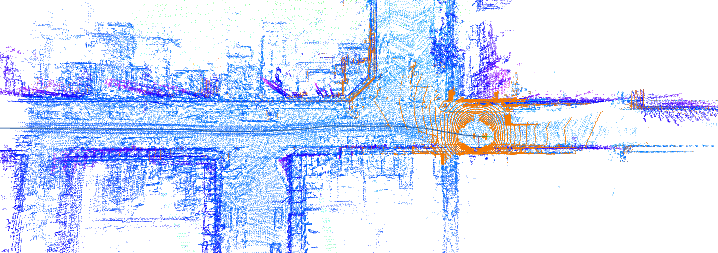}\label{urbanNav}}
    \caption{Odometry estimation results of LIO-EKF with different sensing platforms in different environments.}
    \label{fig:motivation}
    \vspace{-0.5cm}
\end{figure}

Ego-motion estimation is crucial for autonomous robot navigation. The knowledge of the robot's location enables higher-level tasks such as mapping, path planning, or obstacle avoidance. 
In the last decade, 3D LiDAR-inertial odometry (LIO) systems have attracted significant interest in the research community~\cite{xu2021fastlio,li2021liliom,bai2022fasterlio,qin2020lins,shan2020LIOSAM,ye2019liomapping,huang2022ITSM}. Combining LiDAR and IMU measurements provides a low-drift ego-motion estimation, which is robust to aggressive motion profiles, low-light conditions, and poorly structured environments. However, existing LiDAR-inertial systems require parameter tuning to provide an accurate pose estimate \cite{shan2020LIOSAM, li2021liliom, qin2020lins}. In particular, most of the tuning effort is focused on the point cloud registration module, where the feature extraction, number of iterations, and data association threshold must be properly set.

Recently, Vizzo~\etalcite{vizzo2023ral} proposed KISS-ICP, a LiDAR odometry system based on point-to-point iterative closest point (ICP) that can accurately estimate the ego-motion of the robot in a variety of scenarios basically without parameter tuning. Although effective, KISS-ICP has the same limitations as most other LiDAR odometry pipelines~\cite{wang2021floam,Zhang2017loam, shan2018LeGOLOAM, dellenbach2022ct}, for example, poor robustness to motion characterized by rapidly-changing acceleration. We tackle this issue by proposing an equivalent, easy-to-use system for LIO.

The main contribution of this paper is a tightly-coupled LIO system based on point-to-point registration and the classical extended Kalman filter (EKF) scheme. Our system builds on top of the insight provided by KISS-ICP and provides a robust and effective ego-motion estimation by fusing LiDAR scans and IMU measurements. In particular, we show that, given the initial pose prediction from the IMU, an accurate pose can be computed by relying on a standard EKF scheme without the need for multiple iterations as in existing systems. Moreover, to reduce the number of parameters that need to be tuned, we design a novel adaptive threshold for data association that considers the uncertainty of the motion prediction coming from the IMU, the map discretization error, and the noise of the range measurements from the LiDAR. In this way, in our system, no tuning is required for feature extraction, data association, and the number of iterations in the correction step. Fig. \ref{fig:motivation} illustrates the localization and mapping results of our approach running on different challenging datasets.

In sum, we make three key claims: LIO-EKF (i) is on par with the state-of-the-art LIO systems in terms of estimation accuracy, (ii) can provide an accurate pose estimate in different environments and vehicle motion profiles, (iii) can provide pose estimates at close-to-IMU frame rate. These claims are backed up by our experimental evaluation. 

\section{Related Works}

A typical LIO system can be divided into two main components: front end and back end~\cite{slamreview2016}. In the front end, the system performs data association to find corresponding points between the current LiDAR scan and the previous scan (or a map)~\cite{guadagnino2022ral,digiammarino2022iros,dellacorte2018icra}. In contrast, the back end utilizes various state estimation methods to fuse information from both LiDAR and IMU sensors to estimate the robot's pose. 

In the front end of LIO systems, feature-based registration is the predominant paradigm. Feature-based registration, exemplified by LOAM~\cite{Zhang2017loam} and its variants~\cite{shan2018LeGOLOAM, lin2019Loamlivox, wang2021floam}, involves extracting edges and planar patches from LiDAR scans and then finding corresponding features in the map to estimate the robot pose. LIO-SAM~\cite{shan2020LIOSAM} builds on the front end of LOAM and optimizes the robot state estimates from LiDAR and IMU using a sliding window based on a factor graph. LIO-Mapping~\cite{ye2019liomapping} uses a similar front end but includes a rotation-constrained refinement method to further enhance pose estimation and point-cloud maps. LINS~\cite{qin2020lins} also relies on edges and planar features for data association. LiLi-OM~\cite{li2021liliom} and FAST-LIO~\cite{xu2021fastlio} tailor a similar feature extraction module for solid-state LiDARs. Although FAST-LIO2~\cite{xu2022TRO} proposes to register the raw points without feature extraction, it relies on the point-to-plane metric, which also needs local plane approximation and normal vector computation. All of the above approaches require parameter tuning for feature extraction, depending on the specific LiDAR sensor in use and the structure of the environment. Conversely, our proposed method removes such a limitation and uses the classical point-to-point metric in the registration, following the insights of KISS-ICP~\cite{vizzo2023ral}.  

In terms of the back end, most LiDAR-inertial odometry systems rely on either the iterated extended Kalman filter (IEKF)~\cite{xu2021fastlio, xu2022TRO} or factor graph optimization~\cite{kaess2012isam2, loeliger2004factorgraph,merfels2016pose}. Although earlier methods fuse LiDAR and IMU measurements using factor graphs~\cite{shan2020LIOSAM,ye2019liomapping,li2021liliom}, recently the IEKF approaches~\cite{qin2020lins,xu2021fastlio,xu2022TRO} are gaining prominence. In an IEKF, the correction step of the filter is performed multiple times to reduce the linearization errors at the cost of increased computation. In our study, we suggest that, in most scenarios, the classical error-state Extended Kalman Filter (EKF) can be employed to fuse LiDAR scans and IMU readings effectively without additional iterations. This is because that we use a proper strapdown INS model~\cite{Shin2005}, which can provide an accurate initial pose estimate between two successive scans.

Overall, our proposed LIO system, LIO-EKF, employs a point-to-point registration in the front end and uses classical EKF without iterations. Although composed of very straightforward components, our approach achieves comparable performance to state-of-the-art methods in pose estimation accuracy while being significantly faster. 

\begin{figure*}
	\centering
	\includegraphics[width=17cm]{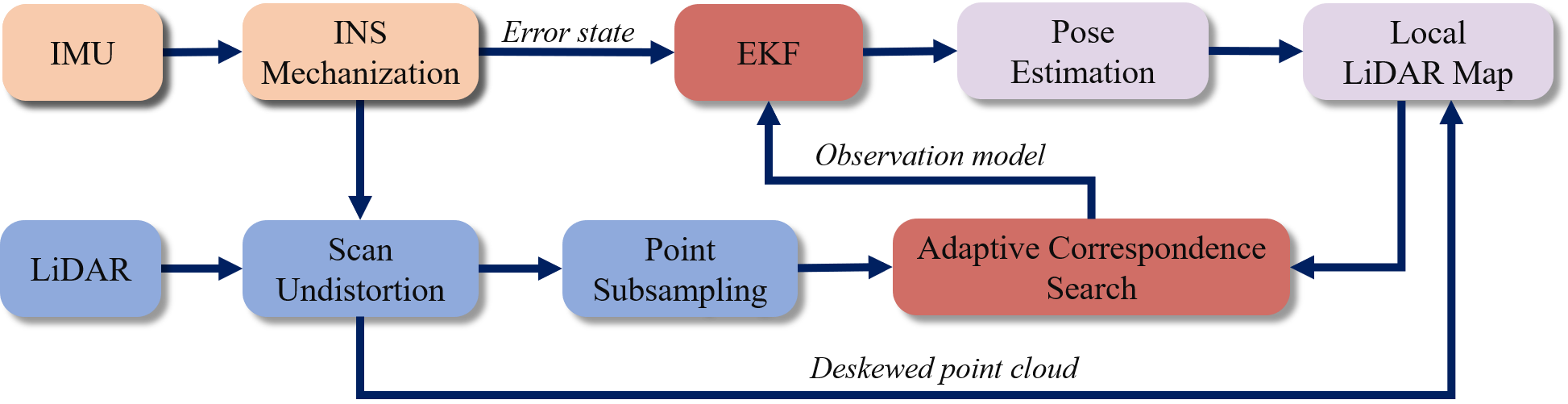} 
	\caption{Overview of LIO-EKF. We integrate the most basic front end (point-to-point association) and back end (EKF) to build a simple, small yet accurate, generic, and high frequency LIO system.}
	\label{fig:overview}
\end{figure*}

\section{Methodology}
We aim to incrementally estimate the pose of a mobile robot by fusing sensor measurements obtained from a LiDAR scanner and an IMU in a tightly-coupled manner via EKF. We employ a conventional strapdown inertial navigation system (INS) for state prediction using the IMU readings. The predicted state is then corrected using the LiDAR scans in an observation model based on point-to-point residuals. To find reliable correspondences in the observation model, we propose an adaptive threshold that takes into account the predicted state uncertainty, map discretization errors, and LiDAR range noise. 

In the following sections, we will first explain the error state model that our system uses. Then, we present the point-to-point based measurement model with associated preprocessing on the incoming LiDAR scan. Lastly, we introduce our novel adaptive thresholding module used for data association in the correction step of the EKF. In Fig. \ref{fig:overview}, we show an overview of our approach.

\subsection{Error State Model}
The robot state can be represented as
\begin{equation}
\bm{x}=\{\bm{t} \ , \ \bm{v} \ , \ \mathbf{R} \ , \ \bm{b}_{g} \ , \ \bm{b}_{a} \} \in \mathbb{R}^{6} \times \mathrm{SO}(3) \times \mathbb{R}^{6},
\label{state}
\end{equation}
where $\bm{t}$, $\bm{v}$, and $\mathbf{R}$ are the position, velocity, and orientation of the robot in the global frame, and $\bm{b}_{g}$, $\bm{b}_{a}$ are the biases of the gyroscope and accelerometer respectively. As new gyroscope and accelerometer measurements are received, we can update the robot's state. We adopt an accurate INS mechanization model~\cite{Shin2005, Groves2013} from the inertial navigation community, as it proves to be much more effective than a vanilla propagation scheme~\cite{shan2020LIOSAM} especially when the vehicle undergoes significant motion. 

Given an accelerometer measurement $\mathbf{a}_k$ and a gyroscope reading $\boldsymbol{\mathbf{\omega}}_{k}$ in robot body frame at time $k$, we can compute the new state $\tilde{\bm{x}}_k$ through motion prediction. We indicate with $\tilde{\mathbf{y}}$ a variable $\mathbf{y}$ that has been updated with the IMU measurements but not with the LiDAR scan. We can write the new state after INS prediction as
\begin{equation}
\begin{aligned}
\tilde{\bm{t}}_{k}&= {\bm{t}}_{k-1} + \frac{1}{2}({\bm{v}}_{k-1}  + {\bm{v}}_{k})\,s_k,
\\
\tilde{\bm{v}}_{k}&= {\bm{v}}_{k-1} + \bm{g}\,s_k + \mathbf{R}_{k-1}\left[\mathbf{a}_k\,s_k + \frac{1}{2}(\boldsymbol{\mathbf{\omega}}_{k} \times \mathbf{a}_k)\,s_k^{2}\right.\\
&\left.+ \frac{1}{12}(\boldsymbol{\mathbf{\omega}}_{k-1} \times \mathbf{a}_k + \boldsymbol{\mathbf{\omega}}_{k} \times \mathbf{a}_{k-1})\,s_k^{2}\right],
\\
\tilde{\mathbf{R}}_k&= \mathbf{R}_{k-1}\text{Exp}(\boldsymbol{\mathbf{\omega}}_k\,s_k + \frac{1}{12}(\boldsymbol{\mathbf{\omega}}_{k-1} \times \boldsymbol{\mathbf{\omega}}_k)\,s_k^{2}), \\
\tilde{\bm{b}}_{g,k}&= \bm{b}_{g,k-1}, \\
\tilde{\bm{b}}_{a,k}&= \bm{b}_{a,k-1},
\end{aligned}
\label{state-dynamics}
\end{equation} 
where $\text{Exp}$ is the exponential mapping to the $\mathrm{SO}(3)$ group, $\bm{g}$ is the gravity vector in the global frame, and $s_k$ is the integration interval. 

To better handle the non-linearities of the state dynamics in ~Eq. (\ref{state-dynamics}), we adopt an error-state formulation of the extended Kalman filter ~\cite{roumeliotis1999circumventing}. In this way, instead of directly estimating position, velocity, attitude, and IMU biases, we model the robot state $\bm{x}$ in terms of the error state vector
\begin{equation}
{\Delta \bm{x}}=\left[\Delta \bm{t}^{\top} \ , \ \Delta \bm{v}^{\top} \ , \ \Delta \bm{\phi}^{\top} \ , \ \Delta \bm{b}_{g}^{\top} \ , \ \Delta \bm{b}_{a}^{\top} \right]^{\top} \in \mathbb{R}^{15},
\label{statevector}
\end{equation}
where $\Delta \bm{t}$, $\Delta \bm{v}$ and $\Delta \bm{\phi}$ indicate the position, velocity, and attitude errors respectively, and $\Delta \bm{b}_{g}$ and $\Delta \bm{b}_{a}$ denote the bias errors of the accelerometer and gyroscope. 
There are several models in the literature describing the time-dependent behavior of the error state in an inertial system~\cite{Shin2005}. As we focus on robot ego-motion estimation with low-cost inertial sensors, we do not consider the earth rotation and the variation of the navigation frame. Therefore, a simplified Phi-angle error-state model~\cite{niu2021} can be used. In particular, we use the discretized model
\begin{equation}
\begin{aligned}
\mathbf{f}(\Delta \bm{x}_{k-1}, \mathbf{a}_k, \boldsymbol{\omega}_k) &= \begin{bmatrix}
\Delta \bm{t}_{k-1} + \Delta \bm{v}_{k-1}s_k \\
\Delta \bm{v}_{k-1} + \left(\mathbf{d}_k\!+\!\mathbf{R}_{k-1}\Delta \bm{b}_{a,k-1}\right)s_k\\
\Delta \bm{\phi}_{k-1}-\!{\mathbf{R}_{k-1}}\,\Delta \bm{b}_{g, k-1}s_k \\
\Delta \bm{b}_{g,k-1} \\
\Delta \bm{b}_{a,k-1}
\end{bmatrix},\\
\\
\text{where} \quad\mathbf{d}_k &= \mathbf{R}_{k-1}\mathbf{a}_k\times\Delta \bm{\phi}_{k-1}.
    \label{systemmodel}
\end{aligned}    
\end{equation}

We can then write the prediction step of the filter as
\begin{equation}
    \begin{aligned}
        \Delta \tilde{\bm{x}}_{k} &= \mathbf{f}(\Delta \bm{x}_{k-1}, \mathbf{a}_k, \boldsymbol{\omega}_k),\\
        \tilde{\mathbf{\Sigma}}_{k} &= \mathbf{A}_{k}\,\mathbf{\Sigma}_{k-1}\,\mathbf{A}^{\top}_{k} + \mathbf{B}_{k}\,\mathbf{\Sigma}_{u, k}\,\mathbf{B}^{\top}_{k} + \mathbf{\Sigma}_{b,k},
    \end{aligned}
\end{equation}
where $\tilde{\mathbf{\Sigma}}_{k}\in\mathbb{R}^{15\times15}$ is the error state covariance, $\mathbf{\Sigma}_{u, k}\in\mathbb{R}^{6\times6}$ is the measurement noise of accelerometer and gyroscope, $\mathbf{\Sigma}_{b,k}\in\mathbb{R}^{15\times15}$ is the process uncertainty including IMU biases noise, and $\mathbf{A}_{k}\in\mathbb{R}^{15\times15}$ and $\mathbf{B}_{k}\in\mathbb{R}^{15\times6}$ are Jacobians defined as
\begin{equation}
    \mathbf{A}_k = \frac{\partial \mathbf{f}(\Delta \bm{x}_{k-1}, \mathbf{a}_k, \boldsymbol{\omega}_k)}{\partial \Delta \bm{x}_{k-1}}, \quad \mathbf{B}_k = \frac{\partial \mathbf{f}(\Delta \bm{x}_{k-1}, \mathbf{a}_k, \boldsymbol{\omega}_k)}{\partial (\mathbf{a}_k, \boldsymbol{\omega}_k)}.
\end{equation}

\subsection{LiDAR Observation Model}
Every time we process an incoming scan, we first deskew the point cloud of the scan using the IMU predicted pose. After scan deskewing, we employ the sub-sampling strategy proposed in KISS-ICP~\cite{vizzo2023ral} to reduce the number of 3D points in the LiDAR point cloud and construct a local map using a voxel grid. We then transform the downsampled LiDAR frame to the robot body frame via the extrinsic calibration matrix between LiDAR and IMU. Please refer to \mbox{Vizzo~\etalcite{vizzo2023ral}} for the details of map maintenance and update. The correspondences between the current downsampled LiDAR frame and the local map are computed via the nearest neighbor search using voxel hashing, resulting in a correspondence set:
\begin{equation}
    \mathcal{C} = \{\,(i,j) \,:\, \min_{i,j} \|\bm{p}_i - \bm{q}_j\|_{2} < \tau\},
    \label{data-association}
\end{equation}
where $\bm{p}_i$ is the $i$-th point in the downsampled LiDAR cloud, $\bm{q}_j$ is the $j$-th point in the local map, and $\tau$ is the maximum correspondence distance allowed for an inlier association.

The observation model for the $i$-th measured point $\bm{p}_{i}$ is:
\begin{equation}
\mathbf{h}_{i}(\Delta \tilde{\bm{x}}_{k}) = \text{Exp}(\Delta \tilde{\bm{\phi}}_k)\, \tilde{\mathbf{R}}_{k}\, \bm{p}_{i} + \tilde{\bm{t}}_{k} + \Delta \tilde{\bm{t}}_k,
\label{obs_model}
\end{equation}
where $\tilde{\mathbf{R}}_k, \tilde{\bm{t}}_k$ represent the robot pose in the global frame after integrating the IMU readings between the previous and current LiDAR scan using~Eq. (\ref{state-dynamics}). The corresponding innovation $\bm{e}_{ij}(\Delta \tilde{\bm{x}}_k)$ for~$\bm{p}_{i}$ in the scan and corresponding point $\bm{q}_j$ in the map is:
\begin{equation}
    \bm{e}_{ij}(\Delta \tilde{\bm{x}}_{k}) = \mathbf{h}_{i}(\Delta \tilde{\bm{x}}_{k}) - \bm{q}_{j}.
\end{equation}
The Jacobian $\mathbf{J}_{ij}(\Delta \tilde{\bm{x}}_{k})$ of $\bm{e}_{ij}(\Delta \tilde{\bm{x}}_{k})$ is:
\begin{equation}
\mathbf{J}_{ij}(\Delta \tilde{\bm{x}}_{k}) = \begin{bmatrix}
    \mathbf{I} & \mathbf{0} & \left[\tilde{\mathbf{R}}_k\,\bm{p}_{i}\right]_{\times} & \mathbf{0} & \mathbf{0}
\end{bmatrix}\in \mathbb{R}^{3\times15}.
\end{equation}

We can then update the error state vector and corresponding covariance using the alternative Kalman gain formulation proposed by Xu~\etalcite{xu2021fastlio}. In practice, to avoid explicitly forming the Kalman gain matrix, we rearrange the terms of the Kalman update equation as  
\begin{align}
    \Delta \bm{x}_{k} &= \Delta \tilde{\bm{x}}_{k} + \left(\mathbf{H}(\Delta \tilde{\bm{x}}_{k}) + \tilde{\mathbf{\Sigma}}^{-1}_{k}\right)^{-1}\,\mathbf{b}(\Delta \tilde{\bm{x}}_{k}), \\
    \mathbf{\Sigma}_{k} &= \left(\mathbf{I} - \left(\mathbf{H}(\Delta \tilde{\bm{x}}_{k}) + \tilde{\mathbf{\Sigma}}^{-1}_{k}\right)^{-1}\,\mathbf{H}(\Delta \tilde{\bm{x}}_{k})\right)\,\tilde{\mathbf{\Sigma}}_{k},
\end{align}
where
\begin{align}
    \mathbf{H}(\Delta \tilde{\bm{x}}_{k}) &= \sum_{i,j \in \mathcal{C}} \mathbf{J}^{\top}_{ij}(\Delta \tilde{\bm{x}}_{k})\, \mathbf{\Sigma}_{p}^{-1}\,\mathbf{J}_{ij}(\Delta \tilde{\bm{x}}_{k}), \\
    \mathbf{b}(\Delta \tilde{\bm{x}}_{k}) &= \sum_{i,j \in \mathcal{C}} \mathbf{J}^{\top}_{ij}(\Delta \tilde{\bm{x}}_{k})\, \mathbf{\Sigma}_{p}^{-1}\, \bm{e}_{ij}(\Delta \tilde{\bm{x}}_{k}),
\end{align}
and $\mathbf{\Sigma}_{p}$ is the observation noise covariance. In this way, we avoid storing and multiplying the complete Jacobian $\mathbf{J}(\Delta \tilde{\bm{x}}_{k})\in\mathbb{R}^{3n\times15}$ of the $n$ scan points.

Once the updated error state vector has been computed via Eq. (\ref{systemmodel}), the position, velocity, attitude, and biases at the current timestamp $k$ are computed via:
\begin{equation}
\begin{aligned}
    \bm{t}_{k} &= \tilde{\bm{t}}_{k} + \Delta \bm{t}_{k}, \\
    \bm{v}_{k} &= \tilde{\bm{v}}_{k} + \Delta \bm{v}_{k}, \\
    \mathbf{R}_{k} &= \text{Exp}(\Delta \boldsymbol{\phi}_{k})\,\tilde{\mathbf{R}}_{k}, \\
    \bm{b}_{a, k} &= \tilde{\bm{b}}_{a, k} + \Delta \bm{b}_{a,k}, \\
    \bm{b}_{g, k} &= \tilde{\bm{b}}_{g, k} + \Delta \bm{b}_{g, k}.
\end{aligned}
\label{box-plus}
\end{equation}

After that, the error state is set to zero and subsequently estimated when a new LiDAR scan is integrated.

\subsection{Adaptive Data Association Threshold}
In Eq. (\ref{data-association}), we seek to establish associations between LiDAR points and local map points through the nearest neighbor search. To mitigate the effect of outlier correspondences, we introduce a threshold denoted as $\tau$; it limits the maximum allowed distance between corresponding points. Instead of determining this threshold empirically, our approach aims to automatically estimate $\tau$ by considering the uncertainty of the initial pose prediction, map discretization errors, and sensor noise.

\subsubsection{Pose Prediction}
The most critical factor in determining good correspondences is the quality of the pose prediction, namely the distance between the initial estimate and the real pose of the robot. Since, in our case, we compute the pose prediction using IMU measurements, we can compute the uncertainty of the relative pose between two successive scans by integrating the noise from the accelerometer and gyroscope over the time interval between the two LiDAR scans. We follow the methodology of Forster~\etalcite{forster2015manifold} but employ the INS mechanization detailed in Eq. \eqref{state-dynamics}. Consequently, we obtain a pose covariance matrix~$\mathbf{\Sigma}_{r,k}$ that characterizes the uncertainty associated with the relative motion at time $k$. Our next step involves projecting this uncertainty into the space of point-to-point distances generated by the relative motion of the robot. These point-to-point distances can be approximated with an upper bound using the formula proposed by \mbox{KISS-ICP}~\cite{vizzo2023ral}:

\begin{equation}
d(\mathbf{R},\bm{t}) = 2 \ r_\text{max} \sin\Big(\frac{1}{2}\mathbf{\Theta}(\mathbf{R})\Big) +|\bm{t}|_{2}.
\label{distances}
\end{equation}

Here, $\mathbf{\Theta}(\mathbf{R})$ represents the angle corresponding to the rotation matrix $\mathbf{R}$, and $r_\text{max}$ signifies the maximum range of the LiDAR. We can subsequently apply the unscented transform~\cite{julier2000new} in conjunction with~Eq. (\ref{distances}) to obtain the variance $\sigma^2_{\text{p2p}}$ of the point-to-point distances from $\mathbf{\Sigma}_{r,k}$.

\addvspace{0.1cm}

\subsubsection{Map Discretization Errors}
Our pipeline employs a voxel grid as the map representation, with a predetermined maximum number of points $m$ per voxel. This choice speeds up the nearest neighbor search but introduces a discretization error in the voxel grid. We account for this error by modeling a Gaussian over the distances between a scan point and $m$ points uniformly distributed on a surface patch contained in a voxel of size $v$. The variance of this Gaussian can be then computed as:

\begin{equation}
\sigma^2_{\text{map}} = \frac{v^2}{m}.
\end{equation}

\subsubsection{Sensor Noise}
We incorporate sensor noise that affects the range measurements from the LiDAR using the variance $\sigma^2_{\text{range}}$.

\addvspace{0.1cm}

The threshold $\tau$ at time $k$ can then be computed by assuming the three Gaussians to be independent and employing the three-sigma bound:

\begin{equation}
\tau = 3\,\sqrt{\sigma^2_{\text{p2p}} + \sigma^2_{\text{map}} + \sigma^2_{\text{range}}}.
\end{equation}

\addvspace{0.5cm}
\section{Experimental Results}
This work presents LIO-EKF, a LIO system based on EKF, and a point-to-point metric that can provide an accurate pose at a high frequency in different environments without parameter tuning. This section presents real-world experimental results to support our key claims, namely, LIO-EKF (i) is on par with the state-of-the-art LIO systems in terms of estimation accuracy, (ii) can provide an accurate pose estimate in different environments and vehicle motion profiles, (iii) can provide pose estimates at close-to-IMU frame rate. After that, we provide ablation studies to analyze the key characteristics of LIO-EKF.

\subsection{Experimental Setup}

We compare LIO-EKF with two state-of-the-art LIO systems, FAST-LIO2~\cite{xu2022TRO} and LIO-SAM~\cite{shan2020LIOSAM} in three different datasets. We used the default parameters in the open-source code of FAST-LIO2 and LIO-SAM, respectively. We disabled the loop closure module in LIO-SAM for the sake of fairness. The three datasets that we selected are:
\begin{itemize}
    \item \textit{urbanNav}~\cite{hus2023urbanNav} an autonomous driving dataset composed of three sequences recorded in Hong Kong.
    \item \textit{m2dgr}~\cite{yin2021m2dgr} a dataset composed of 7 sequences recorded with a wheeled mobile robot on a university campus in Shanghai. 
    \item \textit{newerCollege}~\cite{ramezani2020newer} a dataset composed of two sequences recorded with a handheld device on the Oxford university campus.
\end{itemize}


The IMUs used in the datasets are consumer-grade micro-electromechanical system (MEMS) IMUs. Please refer to the dateset websites for more details. To evaluate the different methods we use the Absolute Trajectory Error (ATE) \cite{vizzo2023ral} and the KITTI metric~\cite{geiger2012} for the relative error. 

\subsection{System Parameters}
For all the following experiments, we use the same system configuration. The list of tunable parameters of our pipeline are:
\begin{itemize}
    \item Max. points per voxel $m$ 
    \item Voxel size $v$ 
    \item Initial state covariance $\mathbf{\Sigma}_0$ 
    \item Point observation covariance $\mathbf{\Sigma}_p$ 
\end{itemize}

Notice that we do not consider as parameters the noise covariance of the IMU measurements $\mathbf{\Sigma}_{u,k}$, the process covariance $\mathbf{\Sigma}_{b,k}$ related to the biases noise, and the range noise $\mathbf{\sigma}^2_{\text{range}}$ as those can be determined from the sensors datasheets.
\subsection{Odometry Performance}
In the first experiment, we analyze the odometry accuracy of our method and its generalizability to different environments and motion profiles. In Table \ref{table_odometry_comparison}, we present the comparative results of LIO-EKF with the two baselines in all datasets. We average the results for the sequences ``street\_01" to ``street\_05" in m2dgr due to space limitation. Unfortunately, we were not able to run LIO-SAM on the newerCollege dataset because additional attitude estimation output from the IMU is needed to initialize the system.

\begin{table}[t]
    \centering
    \caption{Quantitative Results on the Three Different Datasets}
    \label{table_odometry_comparison}
    \begin{tabular}{cccccc}
        \toprule
        \textbf{Sequence} & \textbf{Method} & \textbf{\makecell{Avg.\\tra.}} & \textbf{\makecell{Avg.\\rot.}} & \textbf{\makecell{ATE.\\tra.}} & \textbf{\makecell{ATE.\\rot.}} \\
        \midrule
        \specialrule{0em}{2pt}{2pt}
        \multirow{3}*{\makecell{urbanNav\\20210517}} & {FAST-LIO2} & $4.11$ & $1.68$ & $\mathbf{17.62}$ & $4.45$ \\
        & {LIO-SAM} & $\mathbf{3.18}$ & $\mathbf{1.40}$ & $20.74$ & $\mathbf{4.20}$ \\
        & {LIO-EKF} & $3.20$ & $1.45$ & $24.73$ & $5.05$ \\
        \midrule
        \specialrule{0em}{2pt}{2pt}
        \multirow{3}*{\makecell{urbanNav\\20210518}} & {FAST-LIO2} & $2.73$ & $1.30$ & $23.02$ & $3.27$ \\
        & {LIO-SAM} & $2.52$ & $1.31$ & $\mathbf{20.37}$ & $\mathbf{2.98}$ \\
        & {LIO-EKF} & $\mathbf{2.20}$ & $\mathbf{1.14}$ & $22.44$ & $7.46$ \\
        \midrule
        \specialrule{0em}{2pt}{2pt}
        \multirow{3}*{\makecell{urbanNav\\20210521}} & {FAST-LIO2} & $3.56$ & $1.63$ & $47.29$ & $5.18$ \\
        & {LIO-SAM} & $\mathbf{2.94}$ & $1.62$ & $\mathbf{30.98}$ & $4.51$ \\
        & {LIO-EKF} & $2.96$ & $\mathbf{1.54}$ & $34.97$ & $\mathbf{4.47}$ \\
        \midrule
        \specialrule{0em}{2pt}{2pt}
        \multirow{3}*{\makecell{m2dgr\\street\_01-05}} & {FAST-LIO2} & $\mathbf{1.62}$ & $\mathbf{0.79}$ & $\mathbf{5.05}$ & ${1.79}$ \\
        & {LIO-SAM} & $3.15$ & $1.52$ & $10.19$ & $4.27$ \\
        & {LIO-EKF} & $1.68$ & $0.83$ & ${5.33}$ & $\mathbf{1.70}$ \\
        \midrule
        \specialrule{0em}{2pt}{2pt}
        \multirow{3}*{\makecell{m2dgr\\street\_06}} & {FAST-LIO2} & $3.41$ & $\mathbf{1.55}$ & $\mathbf{8.93}$ & $2.29$ \\
        & {LIO-SAM} & $3.65$ & $1.65$ & $9.04$ & $2.41$ \\
        & {LIO-EKF} & $\mathbf{3.37}$ & $1.56$ & $9.05$ & $\mathbf{2.27}$ \\
        \midrule
        \specialrule{0em}{2pt}{2pt}
        \multirow{3}*{\makecell{m2dgr\\street\_08}} & {FAST-LIO2} & $\mathbf{1.10}$ & $\mathbf{1.65}$ & $\mathbf{2.12}$ & $\mathbf{1.71}$ \\
        & {LIO-SAM} & $3.73$ & $5.78$ & $4.21$ & $6.36$ \\
        & {LIO-EKF} & $1.28$ & $1.85$ & $2.22$ & $1.88$ \\
        \midrule
        \specialrule{0em}{2pt}{2pt}
        \multirow{2}*{\makecell{newerCollege\\short\_exp}} & {FAST-LIO2} & $1.05$ & $1.01$ & $5.14$ & $2.49$ \\
        & {LIO-EKF} & $\mathbf{0.63}$ & $\mathbf{0.73}$ & $\mathbf{4.16}$ & $\mathbf{1.75}$ \\
        \midrule
        \specialrule{0em}{2pt}{2pt}
        \multirow{2}*{\makecell{newerCollege\\long\_exp}} & {FAST-LIO2} & $1.09$ & $1.33$ & $6.33$ & $4.22$ \\
        & {LIO-EKF} & $\mathbf{0.74}$ & $\mathbf{0.91}$ & $\mathbf{5.14}$ & $\mathbf{2.34}$ \\
        \bottomrule
    \end{tabular}   
    \begin{tablenotes}   
        \footnotesize            
        \item Avg. tra. and Avg. rot. denote the relative translation error (\%) and relative rotation error (deg/m) using KITTI~\cite{geiger2012} metrics, respectively.\\
        \item ATE. tra. and ATE. rot. denote the absolute rotation error (deg) and absolute translation error (m)~\cite{vizzo2023ral}, respectively.\\
        In bold the best performance.\\
        
    \end{tablenotes}
    \vspace{-0.5cm}
\end{table}

The quantitative results show that the proposed approach performs on par with the state-of-the-art LIO systems. Moreover, it shows better generalization capabilities, as we can see by comparing the performances of LIO-EKF and FAST-LIO2 on the \textit{urbanNav} and \textit{newerCollege} sequences. This illustrates that, with a single configuration, LIO-EKF can handle different motion profiles of the platform and different structures of the environment. Moreover, these results show that a classical EKF scheme is sufficient to fuse IMU and LiDAR data to estimate an accurate robot pose without relying on more complex state estimation schemes such as factor graphs or IEKF. This evaluation supports the first two claims made in this paper. 




\begin{table}[t]
	\centering
	\caption{Comparison of the Average Processing Time [ms]}
	\label{table_timing_comparison}
\begin{tabular}{cccc}
    \toprule
     & urbanNav & m2dgr & newerCollege \\
    \midrule
    \specialrule{0em}{2pt}{2pt}
    LIO-SAM & $41.68$ & $60.01$ & - \\
    \specialrule{0em}{1pt}{1pt}
    FAST-LIO2 & $24.03$ & $29.96$ & $9.42$ \\
    \specialrule{0em}{1pt}{1pt}
    LIO-EKF & $\mathbf{12.37}$ & $\mathbf{13.65}$ & $\mathbf{7.36}$ \\
    \bottomrule
\end{tabular}
 \begin{tablenotes}   
        \footnotesize            
\item Average processing time for the scan processing in all the selected datasets. The numbers are reported in milliseconds [ms].
    \end{tablenotes}
\end{table}

\subsection{Computation Efficiency}
In this section, we analyze the computational speed of our method. The results support our third claim that LIO-EKF can compute the robot's ego motion at close-to-IMU frequency. We compute the average processing time of the correction step of the filter once a new LiDAR scan has been received. We chose this because the correction step is the most time-consuming part of the pipeline, including scan preprocessing, data association, and pose update. Furthermore, we also compare the time of both FAST-LIO2 and LIO-SAM in terms of the scan processing in all the selected datasets by averaging the values over the different sequences. All the methods have been run on a desktop machine with an Intel i7-10700 CPU at 2.90~GHz and 32~GB of RAM. The results are shown in~Table \ref{table_timing_comparison}. As we can see from the results, LIO-EKF achieves the fastest processing time for the scan processing, which is close to a regular IMU stream rate (100 Hz). In particular, it is about two times faster than FAST-LIO2 and four times faster than LIO-SAM in most scenarios. This is because our method uses a classical EKF scheme, which is very efficient as the optimization does not require multiple iterations.

\begin{table}[t]
    \centering
    \caption{Comparison of LIO-EKF with Different Number of Iterations}
    \label{table_iteration_comparison}
    \begin{tabular}{cccccc}
        \toprule
        \textbf{Sequence} & \textbf{\makecell{\# iterations}} & \textbf{\makecell{Avg.\\tra.}} & \textbf{\makecell{Avg.\\rot.}} & \textbf{\makecell{Processing \\ Time (ms)}}\\
        \midrule
        \specialrule{0em}{2pt}{2pt}
        \multirow{3}*{\makecell{m2dgr\\street\_05}} & $1$ & $2.64$  & $0.80$ & $11.37$\\
        & $10$ & $2.61$ & $0.77$ & $35.10$ \\
        & $100$ & $2.61$ & $0.76$ & $63.90$\\
        \midrule
        \specialrule{0em}{2pt}{2pt}
        \multirow{3}*{\makecell{newerCollege\\short\_exp}} & $1$ & $0.63$ & $0.73$ & $12.41$\\
        & $10$ & $0.60$ & $0.62$ & $26.92$\\
        & $100$ & $0.55$ & $0.58$ & $84.21$\\
        \bottomrule
    \end{tabular}   
    \begin{tablenotes}   
        \footnotesize     
        \item Avg. tra. and Avg. rot. denote the relative translation error (\%) and relative rotation error (deg/m) using KITTI metrics, respectively.
    \end{tablenotes}
\end{table}

\subsection{Ablation Studies}
In this section, we perform two ablation studies to give better insights into the design choices of our pipeline. In particular, we show the impact of using the EKF scheme compared to an IEKF with different numbers of iterations. Furthermore, we showcase that our novel adaptive data association threshold significantly impacts the generalization capabilities of our proposed LIO pipeline.

\subsubsection{Impact of multiple iterations of the odometry accuracy}

We now investigate if multiple iterations can improve the odometry estimate of LIO-EKF. We replaced the used EKF scheme in our system with an IEKF, where we selected 10 and 100 max iterations for comparison. We use the ``street\_05" sequence of \textit{m2dgr} and the ``short\_exp" sequence of \textit{newerCollege} for this ablation study. We compare the relative error according to the KITTI metric and the average processing time of the scan. The results are shown in Table \ref{table_iteration_comparison}.

As we can see from Table \ref{table_iteration_comparison}, there is no significant improvement in the odometry accuracy with more iterations, with a maximum gap of $0.08\%$ in relative translation error at $100$ iterations in ``short\_exp". Conversely, the processing time is dramatically increased by a factor of $8$. The reasons are two-fold. First, IMU has excellent short-term stability, and it can provide an accurate pose prediction for every LiDAR scan so that the estimation can converge to an accurate pose after one update. Second, our error state formulation better handles the nonlinearity of the system dynamics. This shows that our design decision to use a classical EKF effectively computes the robot pose without relying on more complex state estimation schemes.  
\subsubsection{Adaptive Threshold}
In the second ablation study, we show the generalization capabilities and effectiveness of the proposed adaptive threshold model. We compare fixed threshold settings and the adaptive threshold scheme proposed in KISS-ICP~\cite{vizzo2023ral}. For this comparative analysis, we use two \textit{urbanNav} sequences and two \textit{m2dgr} sequences. In this way, we aim to showcase that the proposed adaptive threshold can be generalized to different motion profiles. Table \ref{table_threshold_comparison} lists the results. 

\begin{table}[t]
	\centering
	\caption{Comparison of LIO-EKF with Different Threshold}
	\label{table_threshold_comparison}
	\begin{tabular}{cccccc}
		\toprule
 		\textbf{Sequence} & \makecell{\textbf{Threshold} {(m)}} & \textbf{\makecell{Avg.\\tra.}} & \textbf{\makecell{Avg.\\rot.}} \\
		\midrule
		\specialrule{0em}{2pt}{2pt}
		\multirow{4}*{\makecell{urbanNav\\20210518}} & {$0.3$} & {$2.23$}  & {$1.25$} \\
		& {$1$} & {$\mathbf{2.18}$}   &  {$\mathbf{1.12}$}   \\
		& {KISS-ICP} & {$2.24$}   & {$1.14$}  \\
		& {Ours} & {$2.20$}  & {$1.14$} \\
		\midrule
		\specialrule{0em}{2pt}{2pt}
		\multirow{4}*{\makecell{urbanNav\\20210521}} & {$0.3$} & {$2.96$}  & {$1.72$} \\
		& {$1$} & {$2.95$}   &  {$1.55$}   \\
		& {KISS-ICP} & {$\mathbf{2.95}$}   & {$1.59$}  \\
		& {Ours} & {$2.96$}  & {$\mathbf{1.54}$} \\
		\midrule
		\specialrule{0em}{2pt}{2pt}
		\multirow{4}*{\makecell{m2dgr\\street\_01}}   & {$0.3$} & {$4.25$}  & {$2.26$} \\
		& {$1$} & {$1.46$}   &  {$0.86$}   \\
		& {KISS-ICP} & {$1.79$}   & {$1.04$}  \\
		& {Ours} & {$\mathbf{1.44}$}  & {$\mathbf{0.85}$} \\
		\midrule
		\multirow{4}*{\makecell{m2dgr\\street\_08}}   & {$0.3$} & {$2.66$}  & {$2.58$} \\
		& {$1$} & {$1.33$}   &  {$1.93$}   \\
		& {KISS-ICP} & {$2.06$}   & {$2.50$}  \\
		& {Ours} & {$\mathbf{1.28}$}  & {$\mathbf{1.85}$} \\
		\bottomrule
	\end{tabular}   
 	\begin{tablenotes}   
		\footnotesize     
    \item Avg. tra. and Avg. rot. denote the relative translation error (\%) and relative rotation error (deg/m) using KITTI metrics, respectively. In bold the best performing system configuration.
	\end{tablenotes}
\end{table}

We can see in Table \ref{table_threshold_comparison} that the proposed adaptive threshold achieves the best generalization capability overall. To be specific, for \textit{urbanNav}, \SI{1}{\m} seems to be as good as the proposed. However, it is worse in the \textit{m2dgr}. That means a hand-crafted threshold cannot be guaranteed to work well with different vehicle motions in different environments. In addition, KISS-ICP uses historical registrations to estimate the average deviation between the constant velocity motion prediction and the corrected pose to indicate the threshold statistically. However, it is just an approximation of the motion prediction error, which cannot represent the exact uncertainty of the initial pose guess. In the proposed algorithm, we directly use the uncertainty of the IMU-predicted initial pose, which is more accurate than the statistical model proposed in KISS-ICP. Furthermore, we consider the range noise of the LiDAR sensor and the map discretization error, which better models the data association problem.

\section{Conclusions}

This paper presents LIO-EKF, a LiDAR-inertial odometry system based on a classical EKF scheme. Our approach exploits the classical point-to-point metric with an EKF to build a generic, tightly-coupled LIO system. Thanks to our error state formulation, we can better handle the non-linearities of the problem and converge to an accurate robot pose with a single correction step without relying on multiple iterations, as in IEKF schemes. Additionally, we propose a novel adaptive threshold model considering the pose uncertainty, map discretization error, and sensor noise for a more robust and effective data association.

Extensive real-world experiments conducted with different platforms in different environments show that LIO-EKF can achieve on par odometry performance compared to the more sophisticated state-of-the-art systems with a significantly more straightforward system design. Additionally, LIO-EKF can compute the pose at close to the IMU frame rate (100 Hz), which is much faster than other state-of-the-art methods.

\bibliographystyle{IEEE}
\bibliography{lio_ekf_bib}
\end{document}